# Recognizing Handwriting Styles in a Historical Scanned Document Using Unsupervised Fuzzy Clustering

Sriparna Majumdar[1,2], Aaron Brick[1]

[1] Computer Science Department, City College of San Francisco, San Francisco, United States of America

[2] Department of Psychology, Santa Clara University, Santa Clara, United States of America

Correspondence: smajumda@mail.ccsf.edu

**Abstract:**

The forensic attribution of the handwriting in a digitized document to multiple scribes is a challenging problem of high dimensionality. Unique handwriting styles may be dissimilar in a blend of several factors including character size, stroke width, loops, ductus, slant angles, and cursive ligatures. Previous work on labeled data with Hidden Markov models, support vector machines, and semi-supervised recurrent neural networks have provided moderate to high success. In this study, we successfully detect hand shifts in a historical manuscript through fuzzy soft clustering in combination with linear principal component analysis. This advance demonstrates the successful deployment of unsupervised methods for writer attribution of historical documents and forensic document analysis.

**Introduction:**

Handwriting recognition and separating handwriting into different scribes are computationally challenging tasks. Scribe attribution to handwriting is central to the forensic document analysis of new or historical handwritten documents ([1]). This process requires automation to reduce human bias. In the recent times many research groups in computer vision have focused efforts to build digital optical character recognition tools to facilitate handwriting based classification of historical archived documents([2, 3]), offline signature verification([4]) and associate quality of handwriting to visual cognition and motor activities in human subjects of normal, diseased and aged groups([5, 6]). The handwriting recognition problem for historical documents can be broken down into two smaller problems 1) One writer, many languages and 2) Many writers, one language. While the first problem requires some idea of basic units of languages or lexicons, the latter would not require much idea about the language, vocabulary or grammar used, what each character meant, how they should look like and how they should be strung together to satisfy the semantics. Handwritten documents written in a single language can be classified as single writer or multi writers, solely based on how many different handwriting styles are detected. The individual cursive handwriting is often a combination of all the benchmark cursive forms we are aware about ([7]). Two writers would characteristically use variable lengths of connected strokes, would lift pen at different frequencies, have characteristic stroke width pertaining to the quality of the tip of the pen and pressure applied, introduce characteristic slants, loops, tails and other flares to the beginning and end of characters or words, have characteristic height and width of individual characters ([7, 8]).

Many of the earliest works rely heavily on segmentation of words into graphs of characters ([9]). Handwriting features were extracted from the graphs using Hidden Markov model (HMM) ([10]) and other mathematical feature extracting algorithms, like Levenshtein distance analysis that calculates the cost of converting one grapheme into another ([11]), combining stroke width with direction of stroke ([12]), discrete cosine transform that calculates the sum of cosine functions to represent a finite sequence of data points ([13]), cloud of line distribution that yields unique characteristics of characters based on angle information between dominant points ([14, 15]), to name a few. Those features were then classified into scribes using linear or non-linear component reduction and supervised support vector machine algorithms (SVM) ([16, 17]), convolutional neural network (CNN) ([18]) or recurrent neural network (RNN) ([19, 20]). These supervised and semi-supervised methods work well in the writer classification, with up to 97% accuracy in best cases. Refer to Eskenazi et al., Sreeraj and Idicula and Xiong et al. for comprehensive surveys on handwriting features and classification methods in use ([21, 22, 23]). More recent works have used deepAI models like Fragnet-64 and ResNet-20 for classification purposes ([24, 25]). He and Schomaker used feature pyramids and corresponding image fragments to train and test 4 different datasets that yielded upto 97% accuracy ([24]). Chammas et al. combined scale invariant feature transform (SIFT) local descriptors with Vector of Locally Aggregated Descriptors (VLAD) to extract features and principal component analyses (PCA) to select 32 variance-maximized components generated from 128 total feature dimensions ([25]). Using these they got upto 99% accuracy for historical Arabic datasets available at the Balamand library.

Fewer studies have looked into the use of unsupervised clustering to distinguish handwriting. Vuurpijl and Schomaker introduced a cursivity index which compared the number of letters and the number of strokes conjoining the letters ([26]). Cursivity index of 1 would mean one word was written in just 1 pen stroke and all letters in the word touched their immediate neighbors. Using stroke samples they trained a Kohonen Self-Organizing Map neural network which generated 14 features. Using these features they could cluster print, cursive and mixed styles of writing samples into agglomerative hierarchical clusters. Korovai and Marchenko extended this study by introducing curvature and slants of strokes as features ([27]). They used 12 dynamic features and 22 spatial features and k-means algorithms to group handwriting samples into clusters. Christlein et al, ([28]) combined SIFT based unsupervised keypoint localization with grapheme patch matching to assign SIFT-kmeans cluster indices to graphemes. Data that fell in between SIFT-kmeans clusters were filtered out, before generating cluster indices. Those indices were used as targets to train a CNN/ResNet with keypoint image patches as input. For this composite approach, difference of Gaussian (DoG) algorithm was used for blob-detection of SIFT keypoints situated between text lines. PCA was used for dimensionality reduction and maximizing variance.

Although the state of the art methods in writer identification are leaning toward deepAI and semi-supervised learning, handwriting based categorization of historical documents written in primitive language and by many unknown writers would be difficult in absence of a suitable lookup database. Hence to provide an unsupervised alternative, we asked ourselves "Given a set of black & white handwritten documents, can we cut out word-sized pieces and cluster them, to reveal how many different scribes did the writing?" To answer that, we extracted 10 spatial handwriting features from scanned historical manuscripts written in Spanish ([29]) and available freely at the Bancroft Library. The 445 pages long volume "BANC MSS C-A 35" served as our test dataset and is referred hereafter as C-A 35. This volume is part of the 'Archives of California', bibliographic summaries gathered by the research team of H. H. Bancroft for the production of his magisterial 'History of California' series. The documents it contains would be trivial if the originals had not been burned in the fire following the earthquake of 1906. These 'Archives' are now the best remaining approximation of the administrative documents of Mexican and Spanish California.

We have used tesseract to define textual lines and words image bounding boxes (bbox) and scikit-learn, scikit-image, SciPy and OpenCV Python libraries to extract the following features from those bboxes: stroke width as a measure of pen tip used and the pressure applied; connected component, the area of an image bbox containing writing connected by a continuous curved line, as a measure of cursive flow which is dependent on the frequency at which the pen was lifted; corner angle, the angle between corner inked pixel and its neighboring highest intensity center pixel, as a measure of loops in writing and flares seen at the beginning and tail ends of characters and words; orientation of the letters/words/lines as a measure of forward or backward slant; convex area, as a measure of total polygonal image area taken up by each connected component; height, width, aspect ratio distribution of writing in an image bbox, as a measure of character or word size characteristic of each scribe and blobLoG and blobDoG, as statistical measures of the diametric extent of connectivity of constant intensity pixels which are functions of the length of cursive lines. Using these features, linear and non-linear PCA and unsupervised scikit-fuzzy c-means soft clustering algorithm we could successfully cluster three visually distinguishable writing into three distinct scribes with Top-1 100% accuracy in most test cases. In effect this study offers a tool to separate handwriting in an offline document featuring mixtures of scribes or scripts, and measuring similarity between scribes. It remains to be seen whether our method can be used as a universal tool for recognizing handwriting in all languages, and in all document formats.

**Methods:**

**The Data: C-A 35 Document**

The scanned 19$^{th}$ century C-A 35 manuscript is available at the archive.org with unique identifier 168036072_80_17 ([29]). It is a Los Angeles department of state document written in Spanish language and features volumes 5-8 that were written by hand. It shows evidence of 3 visually distinguishable handwriting styles, one each for each volume, and hereafter aliased as "hands". Figure 1 shows the examples of those 3 distinct hands. They are named Hand1, Hand2 and Hand3, based on their order of appearance in the C-A 35 document. The PDF of scanned C-A 35 document was subjected to pdf2image command line tool to batch-convert pages to 200 dpi PNG images ([30]). Detection of lines and word bounding boxes (bboxes) was done using command line tesseract tool ([31]). The bboxes, found in hOCR files, were used as references to crop portions of the PNG images that were subjected to various feature extraction algorithms. In hOCR files, each line bbox is immediately followed by their corresponding set of word bboxes.

The word bboxes were generated from the image within line bbox coordinate and hence they were a subset of the preceding line bbox. Features were extracted from both line and word bboxes. Figure 2A displays an example of a vertical line bbox and corresponding 8 word bboxes derived from the line. Word images are snapshots of random parts of the line image, and may sometimes contain no writing. Such empty word bboxes were auto-eliminated from the feature extraction pipeline.

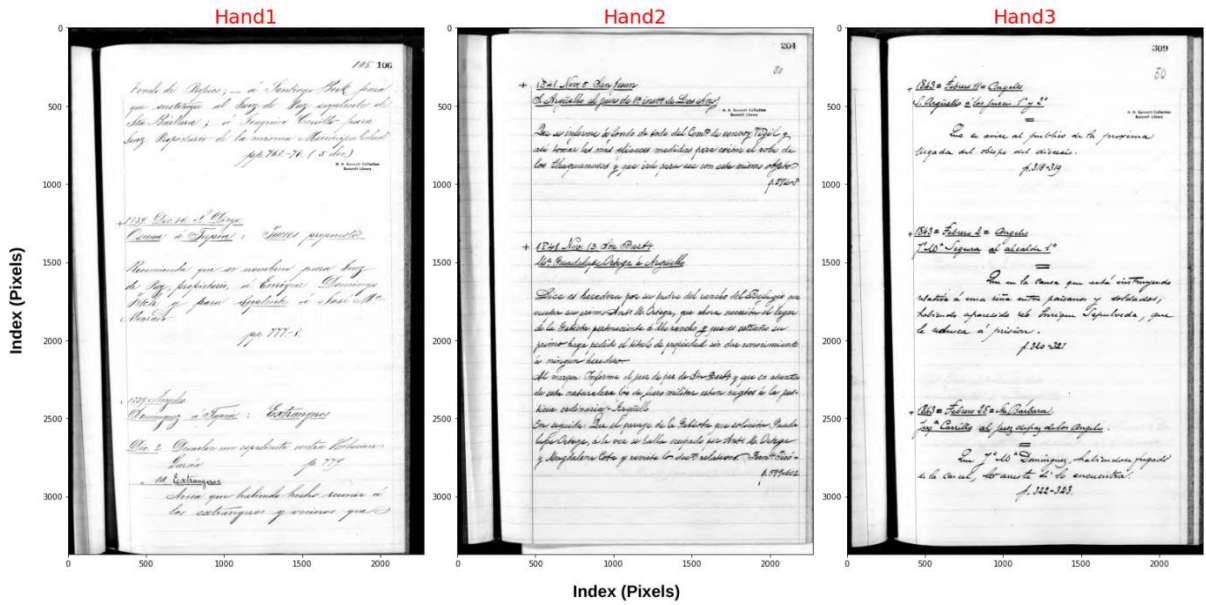

**Figure 1:** *The C-A 35 document features 3 visually distinguishable handwriting styles that we call Hand1, Hand2 and Hand3, serialized in order of their appearance in the document. This figure shows representative pages containing each of these handwriting styles.*

We extracted stroke width (Figure 2B), corner angles (Figure 2C), height and width of characters and words (Figure 2D). The hands with similar size, meaning and look of characters differ in their extent of cursive flow. Our feature extraction algorithms are designed to extract many physical measures of the extent of cursive flow characteristic of each hand, as shown in Figures 2D-F and 3B and discussed in the following sections.

All the feature extracting modules required some preprocessing of the input image. The line and word images were filtered using unsharp-mask, converted to grayscale and inverted to create black background images with white writing before extracting corner angle and stroke width. Height, width measurements using OpenCV and convex area, orientation measurements using `SciPy.ndimage.label` required conversion to grayscale image with white background and black/gray writing. Convex area and orientation extraction additionally required unsharp-mask filtering of the raw image. Extraction of connected component, blobLoG and blobDoG required conversion to grayscale and inversion of the image to create black background image. Stroke width and connected component extraction required further preprocessing: thresholding of the inverted image using OTSU filter, binary image creation and skeletonizing.

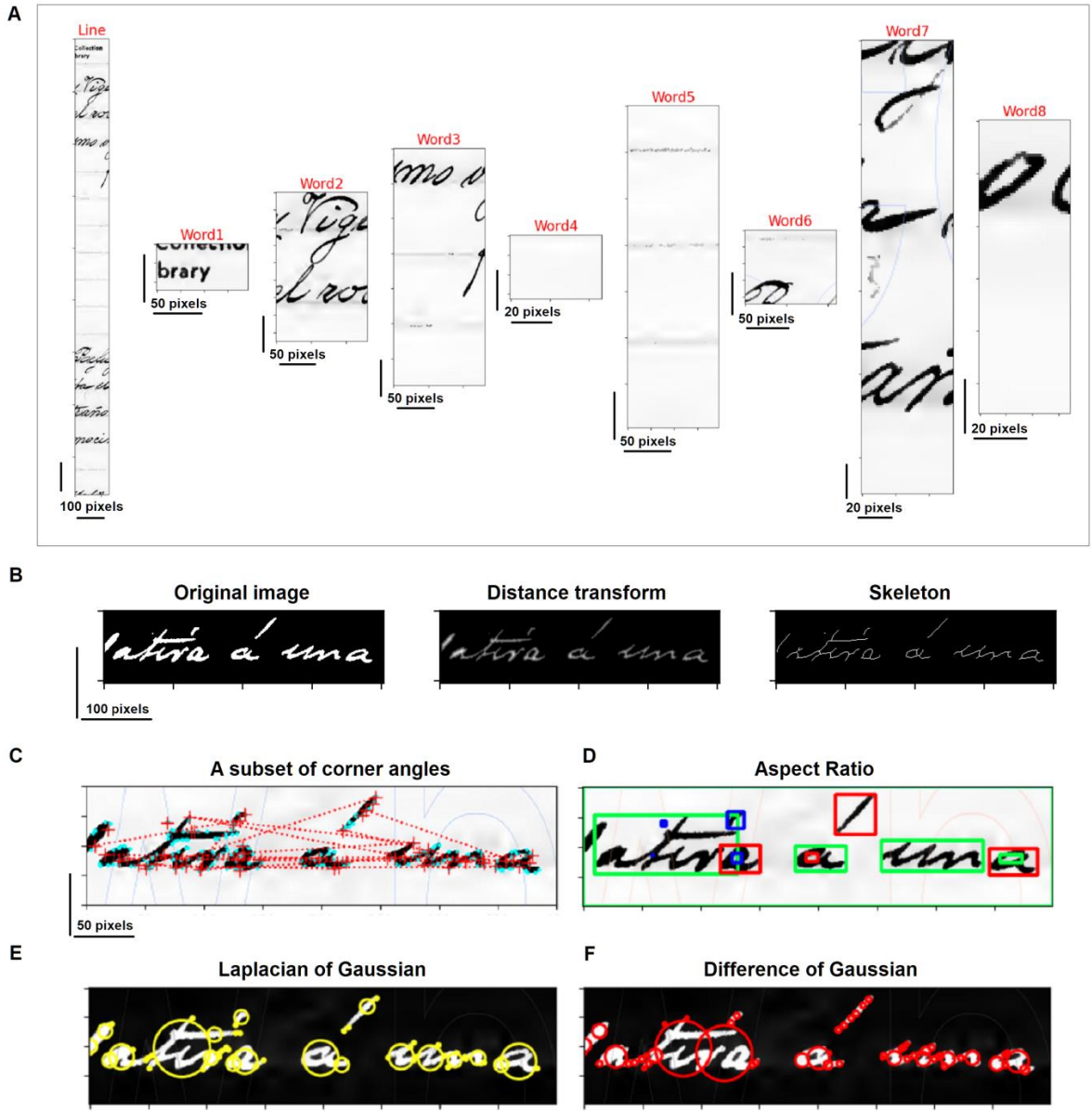

**Figure 2: A.** *A vertical line bbox across Hand2 page, shown in Figure 1, along with word bboxes derived from the line.* **B.** *An inverted word bbox, taken from Hand3 page shown in Figure 1. The pixels in distance transformed image are scaled by their Euclidean distance from the closest 0 valued pixel. The skeleton shows the Boolean mask containing only center pixels. The distance between edge and center were extracted at each point using the skeleton mask, to calculate stroke width.* **C.** *Highest intensity corner peaks are shown in cyan. A subset of them are marked in red '+' and connect to the highest intensity center pixel with red dotted lines. The corner angles are the angles between these lines and the horizontal axis.* **D.** *The cursive graphemes are demarked with color coded boxes. Green: aspect ratio less than 0.5, red: aspect ratio between 0.5-1.0, blue: aspect ratio 1.0 or higher.* **E-F.** *Blobs calculated using blobLoG and blobDoG algorithms.*

The following features have been extracted from the hOCR line and word bboxes of the C-A 35 document:

**Stroke Width ([32, 33, 34, 35])**

The stroke width is defined as the perpendicular distance of the two edges of a pen stroke. We used `distance_transform_edt` of `SciPy.ndimage` to calculate the distance transformed image. Every pixel of the transformed image gave the distance value from its nearest 0 valued pixel. A Boolean mask was created from the skeletonized image that stored True values for center pixels of strokes and False values for everything else. Using this Boolean mask, the distance values for the centers were extracted. Stroke width was calculated as twice the distance of the center pixel from its closest 0 valued pixel. Figure 2B shows an example image, its distance transform image and the skeleton mask.

**Corner Angle ([36, 37])**

The `skimage.feature` method `corner_peaks` was used to calculate corner peaks as highest intensity pixels found in the corners of an image bbox. These corner peaks were compared to the highest intensity center pixel to calculate corner orientation using method `corner_orientations`. The corner orientation, by definition, is the angle (-180 to +180 degrees) of the vector from the corner coordinate to the intensity centroid found in the local neighborhood, calculated using first order central moment. Figure 2C shows an example bbox with corner peaks marked with cyan dots. A subset of corner peaks, denoted by + signs, are connected to the corresponding center peaks. The corner angle is measured as the inclination of these lines with respect to the horizontal axis.

**Convex Area and Orientation ([38, 39, 40])**

`skimage.measure` provides `regionprops` method that returns extensive statistics about an image that is segmented into labeled areas. We used `SciPy.ndimage.label` to group pixels into labels where all pixels with a common label have same integer value and connect to each other either horizontally, vertically or diagonally. For each labeled area `regionprops` extracts the value of convex area, which is the area of a polygon that fits to a particular labeled area. This gives a finite measurement of the total area of individual connected components that is a function of the length of cursive stroke used by the writer. This measurement is similar to the "connected components" feature we discuss later in this section. Orientation of the labeled area is calculated as the angle between the $0^{th}$ axis and the major axis of the ellipse that has the same second order moment as the labeled area. The orientation ranges between -90 and +90 degrees.

**Height, Width, Aspect ratio ([41, 42])**
OpenCV ([41]) provides the cv2 module that is efficient in segmenting images of hands into bboxes of lines, words and characters containing similar values grouped in contours ([40, 41, 42]). We used it to calculate height and width of segmented bboxes of a given image. Height is a good measure of the size of individual characters. We chose height of bboxes that are less or equal to 100 pixels in width, to make sure that we sampled heights of characters or graphemes. Widths of bboxes are representative of character widths, pen strokes used to write words in a cursive flow or extent of a connected line. We used all widths and heights to calculate aspect ratio (width/height) which is a well-recognized characteristic of individual hands ([44]). Figure 2D shows an example bbox with the segmented graphemes represented with colored bboxes around them. Grapheme bboxes are color coded based on their aspect ratio. For all such bboxes, height, width and aspect ratio were extracted as features.

### BlobLoG ([45, 46]) and BlobDoG ([47])

Blobs are regions in an image in which some properties are nearly constant compared to the neighboring regions. Computer vision's blob detection methods can thus be used to identify cursive writing, in which the connected characters would have same stroke width, forward or backward leaning orientation and pixel intensity, properties that can be used to group them in a single blob. Most common blob detectors are based on Laplacian of Gaussian method (LoG) and use scale invariant feature transform algorithm (SIFT, [47]). In this method the image is convolved using a Gaussian kernel for a certain scale to generate a scale-space representation, which is then subjected to a Laplacian operator. The Laplacian is the sum of second order partial derivative of the Gaussian function for each independent variable. In Difference of Gaussian (DoG) approach, the difference of two Gaussian kernel convolutions is computed. LoG can be interpreted as the limiting value of DoG and hence is not too different in their methods of measurements. We extracted diameters of both blobLoG and blobDoG for all hands, to increase the probability of detecting the true extent of a cursive piece of writing. Figures 2E-F display the blobLoG and blobDoG features, presented with circular blobs on an example word bbox.

### Connected Component ([49])

Connected components were calculated using skeletonized images as blobs with "connectivity = 2", meaning diagonally connected components were included alongside horizontally and vertically connected components. `skimage.measure.label` method was used to calculate labels for all blobs. Using those blob labels, `skimage.measure.regionprops` calculated statistics of all segmented regions. The areas of all the regions were stored as the connected components. The difference between features "convex area" and "connected component" is that connected components were calculated using skeletonized images which were segmented and grouped into labels using `skimage.measure.label`, whereas convex area features were extracted from raw grayscale images that were segmented using `SciPy.ndimage.label`. We used two different methods of segmentation, taken from SciPy and scikit-image libraries, to make sure that we included all cursive components, picked by two independent segmentation algorithms. Figure 3 and result section provide further discussion on this feature.

### Standardscaling of data and principal, independent and kernel principal component analyses

One main problem with extracting features from bboxes is that the data in neighboring bboxes are strongly correlated and so do the extracted features. To reduce association between features of neighboring bboxes we used principal (PCA), independent (ICA) and kernel principal (KPCA) component analyses, that project the data into orthogonal spaces and transform the data to maximize the variance in their numeric values. The features require to be scaled suitably for the PCA analyses. Of all the scaling methods provided in `sklearn.preprocessing` module,

`StandardScaler` ([50]) and `MinMaxScaler` ([51]) worked the best. We chose `StandardScaler` for all scaling purposes. `StandardScaler` centers the data on zero mean and unit variance. The standardscaled value z of a feature x is calculated as:

$$z = \frac{(x - a)}{s} \qquad \text{Eq. 1}$$

where a is the mean and s the standard deviation.

The mean and standard deviation for each column are calculated from all samples provided in the dataset and are stored in a one dimensional array with size equal to total number of feature columns. This array is referred to while transforming the data into scaled data.

PCA ([52]) does linear dimensionality reduction by using singular value decomposition (SVD) of the data to project to a lower dimensional space. It centers a multivariate dataset and projects it into sequential orthogonal components with decreasing variance. In scikit-learn PCA works like a transformer, which can learn custom number of components and project the input data onto them. ICA ([53]) can be used to distribute multivariate data in subcomponents that show maximal independence. It works well in separating superimposed signals. KernelPCA ([54]) is the non-linear extension of PCA analyses which achieves dimensionality reduction by use of non-linear kernels. We have used kernel = cosine and kernel = rbf (rbf: radial basis function), as they can work well with angular data and data with implicit periodicity, which we expect to see in features relating to cursive writing. We also tried non-linear dimensionality reduction method t-SNE as computer vision methods use it routinely for visualizing multidimensional data in 2 or 3 dimensions and facilitating with clustering ([55]).

**Scikit-Fuzzy c-means soft Clustering Method ([56, 57])**

Scikit-fuzzy is a Python module originally written by SciPy community and is available on Anaconda platform for Windows 10.0 where we ran all of our feature extraction and data analyses algorithms. Fuzzy logic stems from the idea that there isn't any absolute truth, just a partial truth. This logic is implemented in fuzzy c-means clustering algorithm. In this algorithm, each data point is assigned membership into different neighboring fuzzy clusters, defined by their own calculated mean (c-mean). The Euclidean distance matrix generated by `skfuzzy.cluster.cmeans` method is used to calculate membership of a data in various clusters. Fuzzy partition coefficient (FPC) is a good index for determining the "goodness of fit". FPC is defined as the distance between the fuzzy centers. The closer the FPC value to 1, the better the fuzzy c-means clustering is and the more separated the clusters are. We used FPC values as a reference to accept or reject the algorithm. Fuzzy clusters with FPC values of 0.7 or higher for 2-center fuzzy clustering and 0.6 or higher for 3-center fuzzy clustering were accepted as good fit for our dataset.

**Results:**

**Extracting the features and generating data containing numeric feature rows**

The hOCR files containing bbox coordinates of lines and words and their corresponding PNG images of C-A 35 pages were used to extract 10 features discussed in the methods section. The pixels at the edges of each page were excluded by cropping the images by 5%. This was done to reduce edge related error in feature extraction. Despite of that we often saw the edge effect while parsing individual line or word images. For example, a green bbox with aspect ratio less than 0.5 encompasses the entire image in Figure 2D, raising unusually high height and width values.

Each row of data contained 10 values for each numeric feature, making the total column number 100. The image bboxes that produced less than 5 values for any of the 10 features were excluded, since no reliable statistics can be generated using them. The line bboxes usually produced more than 5 values per feature, whereas word bboxes produced much fewer feature values, often less than 5, causing rejection of several word bboxes. For the hOCR image bboxes producing 5 or more values for a single feature, mean and standard deviation were calculated using the values and were fed to a normalized number generator. The normalized number generator used normal distribution to generate random samples with values in the range (mean ± 2 × standard deviation) for each feature. These random samples were used to make the total number of data points per feature equal to a multiple of 10 using the following formulas where N is the number of extracted feature values, p is the quotient of N and n is the remainder; m is the required number of random samples generated by normalized number generator:

$$n = N - p \times 10 \qquad Eq. 2$$

where $N \geq 5$; $0 \leq p \leq N/10$;

$1 \leq n \leq 9$ when $N \geq 10$, $5 \leq n \leq N$ when $N < 10$

$$n + m = 10 \qquad Eq. 3$$

where $1 \leq m \leq 10 - n$; $(N + m) \bmod 10 = 0$; N, n, m, p are integers

The extracted features were filtered before performing the above calculations, such that N became the total number of original feature values falling in the range (mean ± 2 × standard deviation). This simple Gaussian filtering is required to exclude outliers with huge variance, arising mostly due to edge effect. Cursive features were fewer and required m synthesized data points. m was 0 for most of the non-cursive features. For example, the image in Figure 2B generated 9 connected components, 11 aspect ratios, 43 convex areas, 44 orientations, 55 blobLoG, 56 blobDoG, 215 corner angles and 674 stroke width values. For bboxes like this only connected component would need randomly generated data point to bring up the data size to 10. For the rest of the features 10 random values would be picked to create 1 row of data containing 100 columns.

In case of line bboxes as many rows were generated as possible by serially concatenating 10 values per feature using this formula:

$$S_r = [X_{i,j}] \qquad Eq. 4$$

Where i is the feature index, r is the row index and j is the index of the ith feature in rth row;

$0 \leq i < 10$; $1 \leq r \leq (N + m) / 10$; $0 \leq j < 10$; i, r and j are integers

For each line bbox a loop was iterated to generate r rows of data. The rows were then vertically appended to the dataset. In case of word bboxes only one row was generated, and excess feature values for all features were ignored. This was done to minimize duplication of data, as word bboxes are subsets of preceding line bbox, whose features had already been extracted and included in the preceding rows.

When we followed C-A 35 document page by page, we found a continuous flow of similar looking writing, presumably by a single scribe. There are two title pages across which handwriting shifted from one writer to another: pages 124 and 260. They had completely different character styling, causing their exclusion. They signal real hand shifts in the document. Hand1 spans pages 2 – 123, Hand2 spans pages 125-259 and Hand3 spans pages 261 – 445. As can be seen from the example pages of all 3 hands in Figure 1, Hand1 has big lettering and long descending lines associated with letters like 'p' or 'g' that contain downward shafts. The capital letters also show characteristic flares at their start. Among all the 3 hands, Hand1 appears to have the longest cursive lines. Hand1 writing is sparse, generating 7619 feature rows from first 90 pages averaging at 84.7 rows per page. Hand2 writing is much denser, with somewhat loopy cursive flow and characteristic flares that are distinctly different compared to other two hands. 13601 feature rows were extracted from first 90 pages of Hand2, averaging at 151.1 rows per page. Hand3 is characterized by thicker pen stroke, sparse writing and large fluctuations in the height of individual characters. 7695 feature rows were extracted from first 90 pages of Hand3 averaging at 85.5 feature rows per page.

The features described in the methods are designed to extract spatial information about both graphemes and cursive words. The visualization in Figure 3 presents a graphical description of the handwriting features analyzed. Figure 3A shows a cropped image taken from Hand3 example in Figure 1. The original scanned document shows black writing on white background. The top 2 text lines are not underlined. For these lines 4 bounding lines of the text, the blue dashed lines, help to distinguish capital letters from small letters. Capital letters found in between $1^{st}$ and $3^{rd}$ bounding lines differ in height and width compared to the small letters that are found in between $1^{st}$ and $3^{rd}$, $2^{nd}$ and $3^{rd}$ and $2^{nd}$ and $4^{th}$ bounding lines. The bottom 2 text lines of Figure 3A contain underlines and digits in addition to alphabets. They require definitions of 5 bounding lines. The mutual distances between these 5 bounding lines in the bottom of Figure 3A are evidently different than the top 4 bounding lines. The digits in the underlined texts are found in between $2^{nd}$ and $4^{th}$ bounding lines, capital letters are found in between $1^{st}$ and $4^{th}$ bounding lines and small letters are found in between $1^{st}$ and $4^{th}$, $3^{rd}$ and $4^{th}$ and $3^{rd}$ and $5^{th}$ bounding lines. These differences add to the variability in the height feature as well as features like aspect ratio, convex area and connected component that are dependent on the height of a labeled area on the image. Our height feature represents height of all graphs with width 1-100 pixels. Therefore heights are more representative of single characters, similar to stroke width. Widths of all segmented graphs were included as features; hence width represents the horizontal extent of all segments. blobLoG and blobDoG are composite features, reporting diametric measures of single characters, as well as cursive components along horizontal axis. Similarly aspect ratio, corner angle, orientation and convex area are composite features representing both single characters and words.

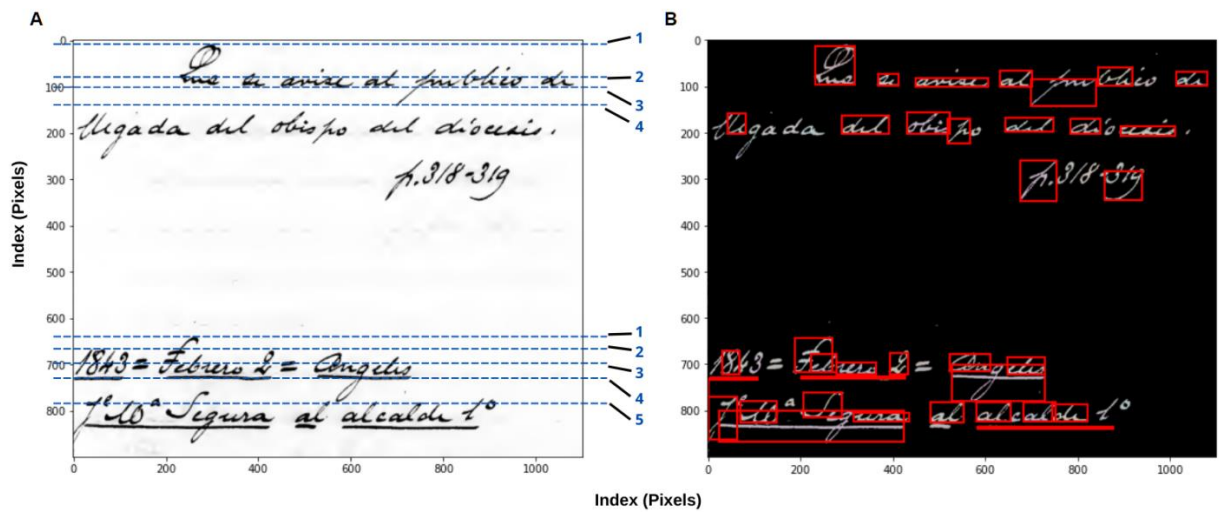

**Figure 3:** *Graphical description of the handwriting features. **A**. A 1100 X 900 pixels cropped image bbox is taken from page 309, Hand3, shown in Figure 1. The text lines toward the top of the image are not underlined. The top 4 bounding lines (blue dashed lines) of the writing contain capital and small letters. The text lines in the bottom of the image are underlined and contain digits, causing the distances between bounding lines change. We require 5 bounding lines to define the areas corresponding to capital letters, small letters and digits. **B**. Measurement of connected component: the image here is the same image shown in **A**, after segmentation by* `skimage.measure.label`. *The red rectangles are the bounding boxes of segmented regions. Segmentation uses skeletonized, inverted, grayscale image as reference, and hence pixels with intensity below threshold are not grouped as segmented regions. The bright pixels within one region have similar intensity values and are touching each other horizontally, vertically or diagonally. The areas of the regions are extracted as connected components.*

Among all features, connected component extraction involves most rigorous image preprocessing (see methods) and typically produces fewer values than width or convex area. All other feature extraction processes require fewer preprocessing steps and report numerous values for a text line (see Figure 2). The red rectangular boxes in Figure 3B mark all the segmented regions of the original image bbox shown in Figure 3A, whose areas were extracted as connected components. The height and width of the red bboxes were calculated using skeletonized image as reference. Hence words or graphemes written with light pen strokes went below threshold and were not demarked with a red rectangular bbox in Figure 3B, as they were not assigned a label.

Width, aspect ratio, convex area, connected component, blobLoG and blobDoG provide direct measures of cursivity and are hence classified here as cursive components. In our datasets features were concatenated in this order: 1. connected component, 2. orientation, 3. height, 4. blobLoG, 5. width, 6. convex area, 7. corner angle, 8. stroke width, 9. aspect ratio, 10. blobDoG.

The features representing cursive components were interspersed with features that represent non-cursive components.

As mentioned earlier, first hand shift in the C-A 35 document happens at page 124 and second at page 260. They are referred here as Handshift1 and Handshift2. We used our feature extraction algorithms to extract features from 2-3 pages at either side of the hand shift points. 65 feature rows for each of Hand1 and Hand2 were picked at Handshift1 and 117 feature rows for each of Hand2 and Hand3 were picked at Handshift2. The title pages (124, 260) were omitted while extracting features. The following section compares hands at these two hand shift points.

**The features for each hand overlap in feature plane at the hand shift points**

The cursive writing is extremely correlated in nature. Stronger correlation is expected to be seen between neighboring word bboxes than neighboring line bboxes. Our data has very large contribution from line bboxes and fewer contributions from word bboxes. When we plotted first

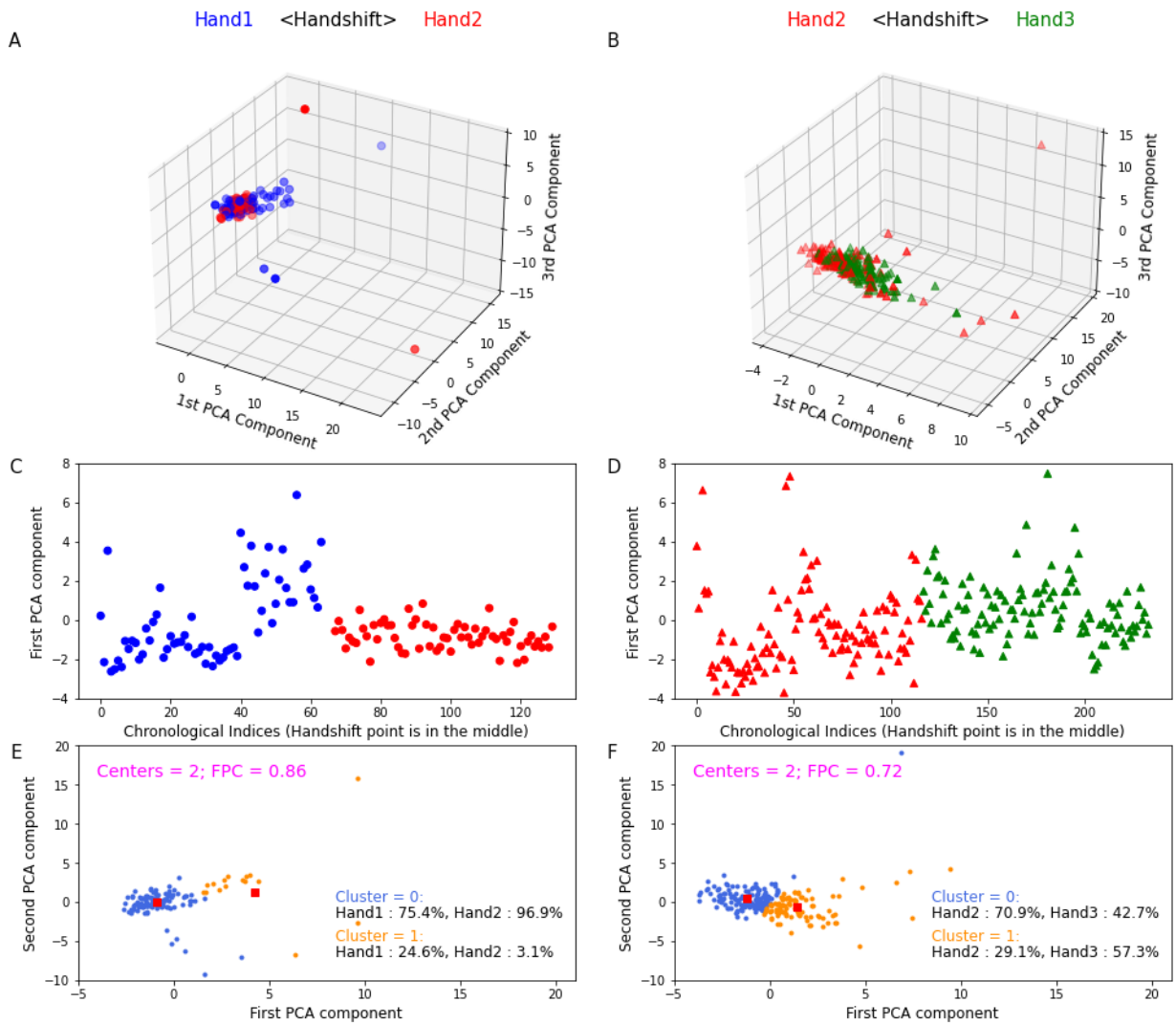

**Figure 4:** *Analyses of writing around two existing hand shift points in the C-A 35 document. The handwriting shifts from Hand1 to Hand2 at page 124 and from Hand2 to Hand3 at*

*260. 65 feature rows per hand were taken around Hand1<Handshift>Hand2 for analyses in **A**, **C**, **E**. 117 feature rows per hand were taken around Hand2<Handshift>Hand3 for analyses in **B**, **D**, **F**. **A** and **B**, 3-dimensional representations of 3 PCA components generated from data around Hand1<Handshift>Hand2 and Hand2<Handshift>Hand3. The hands are color coded as Hand1: blue, Hand2: red, Hand3: green. **C** and **D**, the distribution pattern of first PCA component at either side of the two hand shift points. Spread of data points on the left side of hand shift, with smaller chronological indices than the hand shift point located in the middle, is distinctly different compared to the spread of data points found at the right side of the hand shift point, at higher indices. **E** and **F**, results of 2-center scikit-fuzzy clustering generated using first 2 PCA components of the data around both hand shift points. Scikit-fuzzy differentially distributes the hands from each side of the hand shift points, as indicated by their membership percentages shown in the legend.*

3 PCA reduced components generated from 100 original feature columns, the hands formed distinct yet overlapping clusters in three dimensions. Hand1 and Hand2 data around Handshift1, and Hand2 and Hand3 data around Handshift2 are plotted in Figures 4A and 4B. The hands are color coded as Hand1: blue, Hand2: red and Hand3: green. Figures 4C and 4D show plots of the first PCA component against the chronological indices at the two hand shifts. We saw contrasting patterns of distribution at either side of the hand shift points, which are located at the middle of each plot. Indices lower than the hand shift and indices higher than the hand shift are occupied by data corresponding to two different hands, and hence they are colored differently. Figures 4E and 4F show plots of first PCA component against second PCA component. The hands overlap in these plots, like how they do in Figures 4A and 4B, however the 2-center scikit-fuzzy c-means clustering analyses generated 2 clusters (shown in blue and orange) with distinct cluster centers (red squares) and FPC scores greater than 0.7. The fuzzy memberships of each hand are listed in the legends of Figures 4E and 4F. Evidently each hand has membership in both the fuzzy clusters, although their percent distribution varies significantly, proving that scikit-fuzzy can successfully distinguish between handwriting styles by segregating the PCA components differentially among fuzzy clusters. We also tried hard clustering of hand shifts data using Birch, MeanShift, Agglomerative, Gaussian Mixture and k-means algorithms, where one cluster would be presumably assigned to just one hand ([58, 59]). Those algorithms produced inconsistent clustering results. For example, Birch generated 3 clusters in which all 3 hands showed similar membership. No cluster had majority of data points assigned to just 1 hand, indicating that they were not good enough to cluster our data containing lots of overlaps between the hands, although they produced good results in other studies ([26, 27]). Our features gave good results when supervised RandomForestClassifier was trained with just 100 random feature rows per hand ([60]). Leave-one-out cross validation resulted in Top-1 92% accuracy, with a mean accuracy score of 82% even without any StandardScaling or PCA reduction. This proved our features to be good descriptors of the 3 hands tested.

**Testing efficacy of scikit-fuzzy in terms of distinguishing hands around dummy hand shifts**

We have tried scikit-fuzzy clustering on a big dataset created by serially concatenating first 7600 rows each of Hand1, Hand2 and Hand3, with artificial or "dummy" hand shifts happening at the junctions of hands.

| Serial No. | Datasets | No. of Centers / Fuzzy Partition Coefficients (FPC) | | | |
|---|---|---|---|---|---|
| | | 2 component PCA | 2 component ICA | 2 component KernelPCA (kernel = Cosine) | 2 component KernelPCA (kernel = rbf) |
| 1 | 7600 rows of each: Hand1+Hand2+Hand3 | 2-center FPC = 0.82  3-center FPC = 0.75 | 2-center FPC = 0.83  3-center FPC = 0.76 | - | - |
| 2 | 65 rows each of Hand1 and Hand2 around Handshift1+ 65 rows of Hand3 next to Handshift2 | 2-center FPC = 0.82  3-center FPC = 0.76 | 2-center FPC = 0.81  3-center FPC = 0.80 | 2-center FPC = 0.77  3-center FPC = 0.66 | 2-center FPC = 0.82  3-center FPC = 0.69 |
| 3 | 65 rows of Hand1 next to Handshift1+ 65 rows each of Hand2 and Hand3 around Handshift2 | 2-center FPC = 0.77  3-center FPC = 0.73 | 2-center FPC = 0.76  3-center FPC = 0.70 | 2-center FPC = 0.74  3-center FPC = 0.66 | 2-center FPC = 0.77  3-center FPC = 0.68 |
| 4 | Handshift1 + Handshift2: 65 rows of Hand1, 320 rows of Hand2, 171 rows of Hand3 | 2-center FPC = 0.82  3-center FPC = 0.81 | 2-center FPC = 0.81  3-center FPC = 0.80 | 2-center FPC = 0.77  3-center FPC = 0.64 | 2-center FPC = 0.82  3-center FPC = 0.70 |

**Table 1:** *Four datasets containing various samples of the 3 hands from C-A 35 document were subjected to 2-center and 3-center scikit-fuzzy c-means clustering. The Fuzzy Partition Coefficient is a good measure of the effectiveness of fuzzy distribution of data among distinct clusters. This table shows that in all cases 2-center c-means clustering works better than 3-center clustering.*

Both 2-center and 3-center fuzzy c-means clustering were successful in differentially distributing the 2 component PCA or ICA reduced data in separate clusters, resulting in good separation of cluster centers with high FPC scores (dataset 1, table 1). 2-center clustering showed better separation of the hands with membership percentages differing by 5% or more. For example, 2 component PCA reduction of dataset 1 from table 1 showed the following membership distribution in the 2-center fuzzy clusters: $0^{th}$ cluster: {Hand1: 55.8%, Hand2: 95.4%, Hand3: 44.3%}, $1^{st}$ cluster: {Hand1: 44.2%, Hand2: 4.6%, Hand3: 55.7%}. Evidently, the 3 different hands have unique membership percentages in the 2 clusters that can be used as a signature for each hand. We were unable to try KernelPCA on such huge dataset. KernelPCA algorithm is much slower compared to PCA or ICA, as it performs a non-linear decomposition of features by projecting the features suitably at a lower or higher dimension than the data itself. It could be used successfully for smaller datasets.

The real world handwriting datasets are often quite small, with only a few pages per hand available for feature extraction. Keeping that in mind we created datasets by concatenating data for each hand, picked around pages 124 and 260, the real hand shift points in C-A 35 document. We vertically concatenated equal (datasets 2, 3) or unequal (dataset 4) number of rows for each hand in three different combinations (table 1). 2-center and 3-center scikit-fuzzy could differentially distribute the hands in separate clusters for each data combination, as judged by the FPC scores being always greater than 0.7 for 2-center and always greater than 0.6 for 3-center fuzzy clustering (table 1), with PCA reduction of data helping the clustering the most. Such top-1 100% accuracy in handwriting clustering was not achieved when 2 components t-SNE data reduction algorithm was used instead of PCA or ICA, besides t-SNE turned out to be at least 10 times slower compared to PCA.

To further test the efficacy of scikit-fuzzy in detecting mixture of hands in a given document; we designed dummy hand shift points by putting 2 random pages at either side of a hand shift. We picked 50 consecutive rows for each hand from random locations of the collections containing Hand1, Hand2 and Hand3 data. We called these 50 consecutive rows as 1 page. 2 such random pages of one hand, 100 rows of features in total (left side), were concatenated to other 2 random pages (right side) of either same hand (self<Handshift>self) or a different hand (self<Handshift>other). Hand shift point was defined as the junction of the page 2 of left side and page 1 of right side. In an ideal situation, 2-center scikit-fuzzy should differentially distribute 2 different hands in fuzzy clusters, whereas 2 different samples of the same hand should show almost identical distribution pattern among fuzzy clusters. 2-5 centers fuzzy clustering algorithms were tested for their efficacy in distributing hands among clusters for both self<Handshift>self and self<Handshift>other combinations, as shown in the distribution pattern of the FPC values in Figure 5A. In this plot, summary of 25 random combinations of each of Hand1<Handshift>Hand1 and Hand1<Handshift>Hand2 were compared using first 2 components of PCA, ICA or KernelPCA decomposed data. We saw similar plots for all other self<Handshift>self and self<Handshift>other combinations (not shown). 2-center fuzzy clustering works best for differentially distributing hands in different clusters when tried on 2 PCA components for each dataset, as the mean FPC score is the highest for PCA among all decomposers.

## Higher number of cursive features gives better success rate in detecting real hand shifts

As discussed earlier, if two hands across a hand shift point belong to a single writer as it is in self<Handshift>self combinations, the fuzzy cluster memberships should have been near-identical for both left and right sides. For the statistical measure of membership differences, we set a tolerance score of 5% in case of 2-centers fuzzy c-means clustering. Please note that for more than 2 clusters, tolerance scores need to be lowered to 3-4% as absolute differences in percent membership get smaller. If the membership percentages in the 0[th] cluster of 2-centers fuzzy c-means clusters differed by more than 5% for the left and right side hands of a dummy hand shift point, we called two sides to be "different" and assigned a Boolean difference value to True, designated by non-equality operator != in figures 5B to 5G, assuming that the left side and right side belong to different scribes. Using this paradigm, we calculated the percentage of times left side and right side of a hand shift were different in this way:

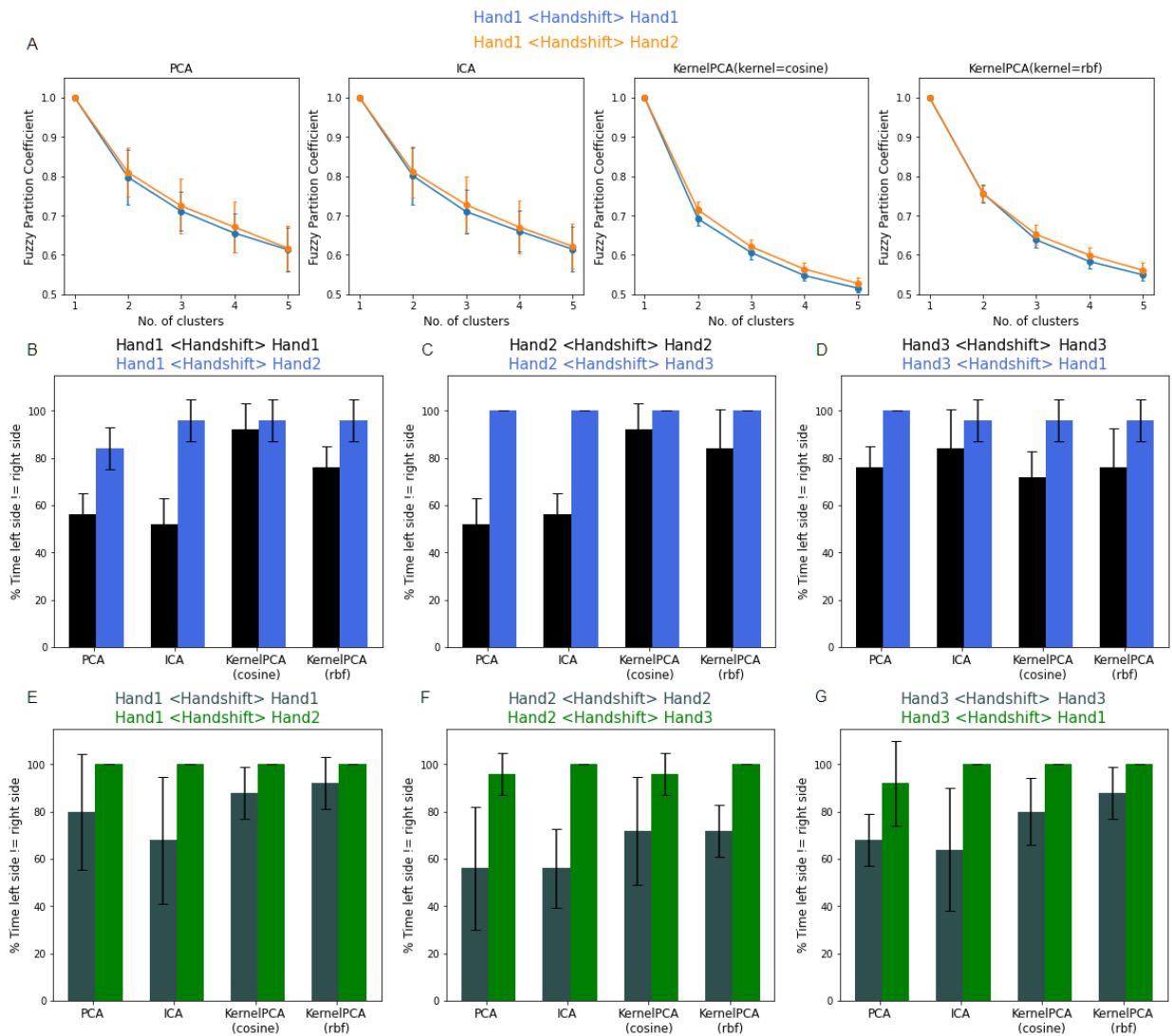

**Figure 5:** *Evaluation of dummy hand shifts. **A**. The plots combine 25 iterations of picking 4 random pages and concatenating them to create 25 different sets of self<Handshift>self and self<Handshift >other dummy hand shifts, where hand shift point is situated at the junction of pages 2 and 3. We used PCA, ICA, KernelPCA (kernel=cosine) or KernelPCA (kernel=rbf) decomposers to reduce 100 columns of StandardScaled features into 2 components for all dummy hand shifts. Four adjacent plots show mean and standard deviation of 25 FPC scores for 2-5 center scikit-fuzzy clustering, in case of each decomposer, as the titles suggest. Hand1<Handshift>Hand1 in blue is compared with Hand1<Handshift>Hand2 in orange. 2-centers scikit-fuzzy c-means clustering works best on PCA components, as the mean FPC scores suggest. Similar result was obtained for Hand2 and Hand3 self<Handshift>self and self<Handshift >other combinations (not shown here).  **B, C, D**, These plots present analyses similar to data presented in Figures **4E** and **4F** which show that 2-center scikit-fuzzy model can differentially cluster the hands from either side of a real hand shift point. We tested whether the membership percentages of a hand at self<Handshift>self dummy hand shift points are similar or different when comparing data from the left side of the hand shift to that taken from the right side. Plots present the mean and standard deviation of percentages of time left and right side memberships differ by more than 5% in 25 iterations. As the histograms suggest the left and right sides for self<Handshift >other are more likely different in their memberships than self<Handshift>self, for all combinations of self and other, with PCA bringing out the most difference among hands. **E, F, G**, These plots present the same analyses done for **B, C, D**, but for 6 selected features (60 columns of features per row). The selected features were orientation, height, width, corner angle, aspect ratio and blobDoG. These features are more related to graphemes and show most difference in their absolute values when compared across different hands. Plots present the mean and standard deviation of percentages of time left and right sides' membership in the cluster 0 of 2-center scikit-fuzzy differed by more than 5%. The data was summarized using 25 iterations. As the histograms suggest the left and right sides for self<Handshift>other are more likely different in their memberships than self<Handshift>self, for all combinations of self and other, with PCA and ICA bringing out the most difference.*

1. We evaluated 5 random dummy hand shifts, with 100 feature rows at each side of the dummy hand shift as described in the previous section, using 2-center scikit-fuzzy clustering of 2 PCA, ICA or KernelPCA components and calculated the Boolean fuzzy membership differences between left and right sides. Less than 5% difference yielded false values, otherwise true.
2. Sum of all 5 Boolean values was divided by 5 and then multiplied by 100 to calculate percentages.
3. 5 such percentages (calculated from 25 total dummy hand shifts) were used to calculate mean and standard deviation.

Data for both self<Handshift>self and self<Handshift>other combinations of hands, generated using above 3 steps are plotted for Hand1, Hand2 and Hand3 in Figures 5B-5D. The plot titles indicate the hand shift combinations tested. It is clear from the histograms that in almost 100 percent of times scikit-fuzzy could assign different membership percentages to two visually different hands found at the opposite sides of the hand shift no matter the type of decomposers used (blue bars in Figures 5B-5D). When data for the same hand was placed at the opposite side of dummy hand shifts scikit-fuzzy assigned differential membership percentages to left and right side at about 60% or less cases when 2 components of PCA or ICA reduced Hand1 or Hand2 data were used (black bars in Figures 5B and 5C). Hand3 features seem to show more variability compared to Hand1 and Hand2, as scikit-fuzzy assigned different membership values for different random pages of Hand3 70-80% of time (black bars in Figure 5D). Hand3 features seem to have higher non-linear components, as KernelPCA with a cosine kernel could extract the similarities among Hand3 pages more often than PCA (Figure 5D). Clearly, scikit-fuzzy c-means clustering is a sensitive method that is good to detect differences among scribes but often raises false positives, in terms of declaring 2 same hand pages as belonging to different scribes. Similar intra-writer variabilities in features were observed in Bengali handwriting recognition ([61]). False negatives were rare in our testing, as seldom were 2 different hands declared as belonging to one scribe.

In an attempt to increase the difference between self<Handshift>self vs self<Handshift>other combinations, with respect to the same hands showing similar fuzzy cluster membership, and distinct hands showing differential fuzzy cluster memberships, we reduced the number of selected features from 10 to 6, by keeping most single character related features and reducing cursive components. We concatenated the selected features in this order: orientation, height, width, corner angle, aspect ratio and blobDoG. Connected component, convex area, blobLoG and stroke widths were omitted, as their absolute values were not too different when compared across 3 hands. This reduced the number of cursive components in feature selected dataset from 6 to 3 and increased the weights of angular features. 2-centers scikit-fuzzy classified the 2 PCA components of feature selected data with an average FPC score of 0.7 or higher. That did not benefit us in our quest to make scikit-fuzzy to assign nearly identical membership percentages to same hands situated across dummy hand shifts. As the plots in figures 5E-5G suggest, the self<Handshift>self combinations (grey bars) show difference in left and right sides more often in case of Hand1 and Hand2 (Figure 5E and 45F), compared to original data with 10 features (black bars in Figure 5B and 5C). We conclude that cursive features contribute more to the differences in handwriting styles in our samples of handwriting, and all 10 features are required to distinguish handwriting. Finally, for many pairwise comparisons, if on average 40-60% times two hand samples turned out to be different, we could reject those differences as false positives and declare them as same hands; conversely, for 2 hands to be different, 2-center scikit-fuzzy should find differential memberships for the 2 hands 80-100% times.

**Gaussian normalization and filtering of outliers help in distinguishing between same vs. different hands**

Our strategy of normalizing and filtering out outliers and having many cursive features gave better separation between same hands vs. different hands pairwise comparisons. In case of data without any normalizing or filtering of outliers, approximately 1000 fewer rows were generated per 100 pages, causing huge loss on data especially from word bboxes. When 6500 unfiltered data rows of each of the 3 hands were serially concatenated, StandardScaled and PCA reduced to 2 components, 2-centers scikit-fuzzy could assign differential cluster membership to them, although the FPC score was lower compared to filtered data (0.73 vs 0.82 in case of filtered data, see Table 1). This also caused less efficient clustering of the hands for small sample size, as was evident from the dummy hand shift comparisons of unfiltered data, when 2 random pages of a hand was compared to 2 other pages of either same or different hands, similar to data presented in Figure 5. Same-hand comparisons resulted differential cluster membership on average 80-90% of times. Different-hand comparisons gave Top-1 100% accuracy.

Table 2 helps in comparing efficacy of 2-center scikit-fuzzy in separating same vs. different hands using the dummy hand shift paradigm for filtered and unfiltered data. For this set of testing we placed just 1 random page per hand across a dummy hand shift, keeping in mind that samples of historical handwriting may sometime contain just 1 or 2 pages. Here are steps that were followed to create dummy hand shifts.

1. At a random start location of a hand dataset 50 consecutive rows were picked as page 1. For page 2 a sliding window skipped 1 row or 50 rows from the start, to select adjacent 50 consecutive rows. Page 1 and 2 were concatenated to form self<Handshift> self dummy hand shift. 50 such slides to the 1-row or 50-rows window were made to create 50 dummy hand shifts.
2. In case of self<Handshift>other comparisons, left and right side of dummy hand shifts were contributed by pages created using step1 applied to 2 different hand datasets. From a random starting location in a dataset, 50 pages per hand were picked where each page differed from the adjacent page by 1 row or 50 rows.
3. The process was iterated 20 times to reset the start location of sliding window to 20 random places in each of the datasets.
4. 2-center scikit-fuzzy clustering was performed on 2 PCA components for each of the 1000 different self<Handshift>self or self<Handshift>other dummy hand shifts created by steps 1-3.

| Page1<Handshift>Page2 2-centers scikit-fuzzy comparison using 2 components PCA. | Unfiltered Data (% time left != right) | | Filtered Data (% time left != right) | |
|---|---|---|---|---|
| | Window width 1 row | Window width 50 rows | Window width 1 row | Window width 50 rows |
| Hand1<Handshift>Hand1 | 66.9±30.5 | 91.8±4.3 | 45.7±23.0 | 66.5±5.8 |
| Hand2<Handshift>Hand2 | 50.3±27.5 | 82.0±8.4 | 40.3±19.2 | 68.0±7.5 |
| Hand3<Handshift>Hand3 | 81.7±11.8 | 93.0±3.8 | 50.6±26.4 | 76.4±7.2 |
| Hand1<Handshift>Hand2 | 94.9±11.7 | 94.9±2.63 | 88.5±24.2 | 91.0±4.6 |
| Hand2<Handshift>Hand3 | 99.0±2.8 | 98.9±1.8 | 94.1±14.3 | 95.7±2.9 |
| Hand3<Handshift>Hand1 | 94.5±10.3 | 97.2±2.4 | 91.5±9.5 | 92.9±3.9 |

Table 2: Sliding window comparison between sample pages of same hands vs. different hands. Each side of the dummy Handshift has 50 rows of either same hands or different hands. Adjacent 50 rows of each hand differed either by 1 row, or 50 rows, based on the width of the sliding window. Normalized filtered data were compared with unfiltered raw data. Higher

*mean values for unfiltered same hand comparisons indicate left and right side of the dummy hand shifts had greater chance of being more than 5% different in their membership in the $0^{th}$ cluster of a 2-center scikit-fuzzy distribution.*

The results from above test are summarized in table 2. The frequency of same hands being categorized as different was significantly higher for unfiltered datasets for all the pair-wise comparisons, as compared to filtered data. Thus normalizing and filtering outliers help in reducing false assignment of same hands to different scribes. Finally, dummy hand shifts testing can be used as a tool for pair-wise comparisons of 2 unknown samples, as many random dummy shifts creation by randomly shuffling and concatenating feature rows, followed by scikit-fuzzy clustering should yield roughly 60% or less difference score for same hands and 80% or more difference score for different hands, making the separation of hands in feature space clearer.

**Discussion:**

Our study intends to detail on 10 handwriting features relating to both graphemes and graphs of words that are written with cursive strokes, demonstrate their effectiveness in numerically defining 3 distinct scribes found in a scanned historical document and show that we can use fuzzy logic to cluster them differentially into more than one cluster. In effect this study offers an unsupervised clustering tool to separate handwriting found in offline documents featuring mixtures of scribes.

In this study we present a way of extracting features using open source image processing tools provided in scikit-learn, scikit-image, SciPy and OpenCV Python libraries accessible on Anaconda platform. We extracted some numeric features related mostly to single characters, like height and stroke width, and some features that were dependent on both single as well as group of characters joined by cursive lines, like width, aspect ratio, diameter of blobs covering one or many characters, area of one or a group of characters, corner angle and orientation. We realized that combining many different measures of cursive flow helped us in extracting the essential characteristics of a scribe. Tesseract was used to detect reference bounding boxes in the scanned images of handwriting. Hence our feature extraction methods relied heavily on hOCR grouping of lines.  Most of the Python libraries that we used in our work segmented the hOCR images further, in order to extract spatial measurements related to characters, words and lines. Five or more cursive components were found in most of the hOCR lines, helping us in getting a reasonable distribution of them. None of our analyses use any information related to the language properties and lexicons. This makes our methods of treating the data language independent.

Our primary goal was to propose a way to cluster scribes in an unsupervised manner. Many past works presented well designed feature extraction algorithms that facilitated supervised classification of labeled handwriting datasets with up to 97% accuracy ([21-23]). Supervised classification would not work for forensic document analytics where none of the handwriting is provided with any label. Our work intended to address that issue by providing a suitable unsupervised method for clustering unknown handwriting found in offline documents. We have shown that unsupervised scikit-fuzzy soft clustering technique can be successfully used in combination with PCA analyses of 10 numeric features, to differentiate between scribes found in C-A 35 document. Fuzzy logic suggests that every hand may belong to more than one cluster. The membership of a data point into a particular cluster depends on the distance of the data point from the cluster center in the feature space. A data point is assigned to the cluster whose cluster center is the closest to it. A hand may have many feature rows that would fall into separate clusters depending on how far they are in the feature space compared to various cluster centers. Each hand is therefore assigned memberships in more than one cluster, based on the distribution of its feature rows among the clusters. We calculated cluster memberships of each hand to show that each hand had unique memberships in fuzzy clusters. We have also shown that 10 features are optimum to bring out the most difference between hands, while making sure that writing by a single scribe are assigned similar cluster memberships by scikit-fuzzy c-means clustering algorithm. Machine learning algorithms provided in scikit-learn suite had been previously used for supervised handwriting recognition purposes ([62]). Our work is one of the first that used scikit's soft clustering algorithms for unsupervised clustering of handwriting. A previous study used soft fuzzy clustering on Persian scripts ([63]). Their approach worked well, although did not reach 100% accuracy. Our features help in frequently reaching 100% accuracy when 1, 2 or many pages of distinctly different hands were compared, although for same hand comparison, false positives were raised approximately half the time. Hence we propose a dummy hand shift paradigm to have a clearer understanding of the differences between 2 previously uncategorized samples (see Table 2).

Our results suggest that KernelPCA decomposition does not work well in bringing out the differences in Hand1 and Hand2 despite the intrinsic nonlinearity in the data. This was surprising for us as we expected KernelPCA to perform better, since other works reported better classification scores for kernelPCA ([10]). This could be due to the fact that angular features don't fluctuate in their numeric values as much as the cursive features. Often the features despite being unique may not strongly differ in their numerical aspects. The cursive features vary greatly in value, with a strong linear trend, compared to stroke width or height of characters. However this does not have to hold true for all handwriting. As we demonstrated in figure 5D, Hand3 features had stronger influence of non-linear components, since KernelPCA worked better compared to PCA. We suggest a comparison of PCA, ICA and KernelPCA components for each handwriting using self<Handshift>self dummy hand shift clustering to come up with the best reducer for each handwriting style before comparing and contrasting with other scribes. Self<Handshift>self can be created using random number of rows of known writer and a known writer can be compared to an unknown writer using many randomized self<Handshift>other combinations of rows. 2-center scikit-fuzzy analyses and tests described in Figure 5 and table 2 can become essential tools to detect real hand shifts using pair-wise comparisons and estimate the number of scribes in a document containing mixture of handwriting, especially benefiting forensic data analytics.